\documentclass[conference]{IEEEtran}
\IEEEoverridecommandlockouts

\usepackage{cite}
\usepackage{multirow}
\usepackage{amsmath,amssymb,amsfonts}
\usepackage{algorithmic}
\usepackage{graphicx}
\usepackage{textcomp}
\usepackage{xcolor}
\usepackage{tabularx}
\usepackage[switch, pagewise]{lineno}
\usepackage[font=small]{caption}
\renewcommand{\thetable}{\arabic{table}}
\def\BibTeX{{\rm B\kern-.05em{\sc i\kern-.025em b}\kern-.08em
    T\kern-.1667em\lower.7ex\hbox{E}\kern-.125emX}}
\begin{document}

\title{Representation Learning on Large Non-Bipartite Transaction Networks using GraphSAGE\\
\thanks{This preprint has not undergone peer review or any post-submission improvements or corrections. The Version of Record of this contribution is published in Graph-Based Representations in Pattern Recognition. GbRPR 2025. Lecture Notes in Computer Science, vol 15727. Springer, Cham., and is available online at https://doi.org/10.1007/978-3-031-94139-9\_17}
}

\author{
\IEEEauthorblockN{1\textsuperscript{st} Mihir Tare}
\IEEEauthorblockA{\textit{NatWest AI Research} \\
\textit{NatWest Group}\\
London, UK \\
mihir.tare@natwest.com}
\and
\IEEEauthorblockN{2\textsuperscript{nd} Clemens Rattasits}
\IEEEauthorblockA{\textit{NatWest AI Research} \\
\textit{NatWest Group}\\
London, UK \\
clemens.rattasits@natwest.com}
\and
\IEEEauthorblockN{3\textsuperscript{rd} Yiming Wu}
\IEEEauthorblockA{\textit{NatWest AI Research} \\
\textit{NatWest Group}\\
London, UK \\
yiming.wu@natwest.com}
\and
\IEEEauthorblockN{4\textsuperscript{th} Euan Wielewski}
\IEEEauthorblockA{\textit{NatWest AI Research} \\
\textit{NatWest Group}\\
Edinburgh, UK \\
euan.wielewski@natwest.com}
}

\maketitle
\thispagestyle{plain}
\pagestyle{plain}

\begin{abstract}
Financial institutions increasingly require scalable tools to analyse complex transactional networks, yet traditional graph embedding methods struggle with dynamic, real-world banking data. This paper demonstrates the practical application of GraphSAGE, an inductive Graph Neural Network framework, to non-bipartite heterogeneous transaction networks within a banking context. Unlike transductive approaches, GraphSAGE scales well to large networks and can generalise to unseen nodes which is critical for institutions working with temporally evolving transactional data. We construct a transaction network using anonymised customer and merchant transactions and train a GraphSAGE model to generate node embeddings. Our exploratory work on the embeddings reveals interpretable clusters aligned with geographic and demographic attributes. Additionally, we illustrate their utility in downstream classification tasks by applying them to a money mule detection model where using these embeddings improves the prioritisation of high-risk accounts. Beyond fraud detection, our work highlights the adaptability of this framework to banking-scale networks, emphasising its inductive capability, scalability, and interpretability. This study provides a blueprint for financial organisations to harness graph machine learning for actionable insights in transactional ecosystems.
\end{abstract}


\section{Introduction}
Graph embedding methods have revolutionised the analysis of networks by providing a way to transform complex network information into low-dimensional vector representations. These embeddings capture the structural, relational, and, when available, feature properties of nodes, making them a powerful tool for applications like fraud detection on financial transaction networks. A taxonomy of graph embedding methods points to three main categories: matrix factorisation, random walk, and GNN (Graph Neural Network) based methods. Matrix factorisation approaches like LINE \cite{b1} and HOPE \cite{b2} focus on approximating adjacency matrices or similarity matrices to embed nodes. LINE preserves both first order and second order proximities, capturing direct connections and shared neighbourhoods, while HOPE extends this approach to high-order proximities, making it particularly effective for embedding directed graphs. These methods have shown promise in tasks such as link prediction and anomaly detection \cite{b3}. However, these methods struggle with large dynamic networks (as observed in financial transaction networks) – they are transductive, meaning they cannot be used to perform inference on unseen nodes without retraining, and have a large computational cost due to the need for matrix decomposition. DeepWalk, introduced by Perozzi et al.\cite{b4}, was a pioneering work that applied truncated random walks to generate node sequences, treating them as sentences in a text corpus and training node embeddings using the Word2Vec model. This method captures community structures effectively and has been applied in fraud detection problems \cite{b5}\cite{b6} where the relationships between nodes are important. Building on this, Node2Vec \cite{b7} improved upon DeepWalk by introducing biased random walks. This allowed the method to interpolate between breadth-first (BFS) and depth-first (DFS) search strategies, providing flexibility to capture community and structural roles within the same embedding framework. Random walk-based approaches have since been studied further with the introduction of metapath2vec by Dong et al. \cite{b12} extending the idea to heterogeneous networks. However, these methods require knowledge of the entire graph during training and are also inherently transductive, hence limiting their scalability to large-scale graphs. Simultaneously, GNNs have emerged as a robust alternative leveraging the power of deep learning to learn embeddings by aggregating information from a node's neighbours. Graph Convolutional Networks (GCNs), proposed by Kipf and Welling \cite{b8}, introduced spectral convolutions for semi-supervised node classification. This method effectively leverages node features alongside graph topology. Similarly, Graph Attention Networks (GATs) \cite{b9} extended GCNs by introducing an attention mechanism to learn the relative importance of neighbouring nodes. In general, GNNs can overcome the transductive limitations of earlier methods by enabling inductive learning, which is necessary for obtaining the node embeddings of dynamic graphs. However, both GCNs and GATs have limitations too as they also require the entire graph to compute the embeddings, causing them to struggle with the same scalability bottleneck on large graphs.

GraphSAGE \cite{b10} addresses these limitations, making it well-suited for large dynamic graphs. GraphSAGE is an inductive representation learning algorithm for graphs, allowing it to infer embeddings for unseen nodes by aggregating information from their local neighbourhoods. This is particularly valuable when working with transaction networks in finance, where new accounts and transactions appear frequently. Moreover, the neighbourhood sampling and aggregation strategies used by GraphSAGE ensure computational efficiency, even when working with graphs containing tens or hundreds of millions of nodes and edges.

Our work builds on the papers by Bruss et al. (2019) \cite{b11} and Van Belle et al. \cite{b13}. The first introduced a financial transaction embedding framework known as DeepTrax using a variant of the metapath2vec model. DeepTrax was used to compute embeddings for merchants on a credit card (point of sale) transaction network and showcased their use in a downstream classification task. Their algorithm uses a random walk-based approach to capture similarities between merchant nodes based on the customers they share. While this approach was effective at dealing with the computational challenge posed by large-scale graph embeddings and capturing ‘semantic’ similarities between the merchant nodes, it also posed some potential limitations. Firstly, their approach relies on a transductive training algorithm that requires the entire graph to be available during training and cannot perform inference on previously unseen nodes, making it less suitable for dynamic financial graphs where new accounts and transactions emerge continuously. Moreover, their approach requires the definition of a metapath, i.e., the pattern of random walks to embed on the graph, which is much harder to define when working with complex non-bipartite financial transaction networks containing multiple node and edge types. Van Belle et al. \cite{b13} addressed the first limitation by applying GraphSAGE, an inductive representation learning algorithm, to a similar tri-partite credit card transaction network to compute embeddings for transactions. However, they utilise a supervised approach to compute embeddings that were trained by using a binary cross entropy loss function on fraud labels. Drawing inspiration from these two works, we use GraphSAGE to compute embeddings for a more complex, bank-wide non-bipartite heterogeneous transactions network. With the aim of forming a holistic view of account behaviours, we extend the graph schema to include account-to-account bank transfers alongside point of sales transactions with merchants. Further, we use an unsupervised loss function to develop embeddings that encapsulate information about accounts' transaction neighbourhoods without bias towards any specific downstream task, resulting in a reusable asset that could be used for multiple downstream tasks. We demonstrate the correctness and validity of our unsupervised embeddings using various quantitative and qualitative methods, and outline the results from their application to a money mule detection task.

\section{Methodology}
In this section, we outline how the graph is constructed, how it is used to train a GraphSAGE model, and how the hyperparameters are tuned.
\subsection{Graph Schema – Population Design}\label{GS}
\begin{figure}[htbp]
\centerline{\includegraphics[scale=0.10]{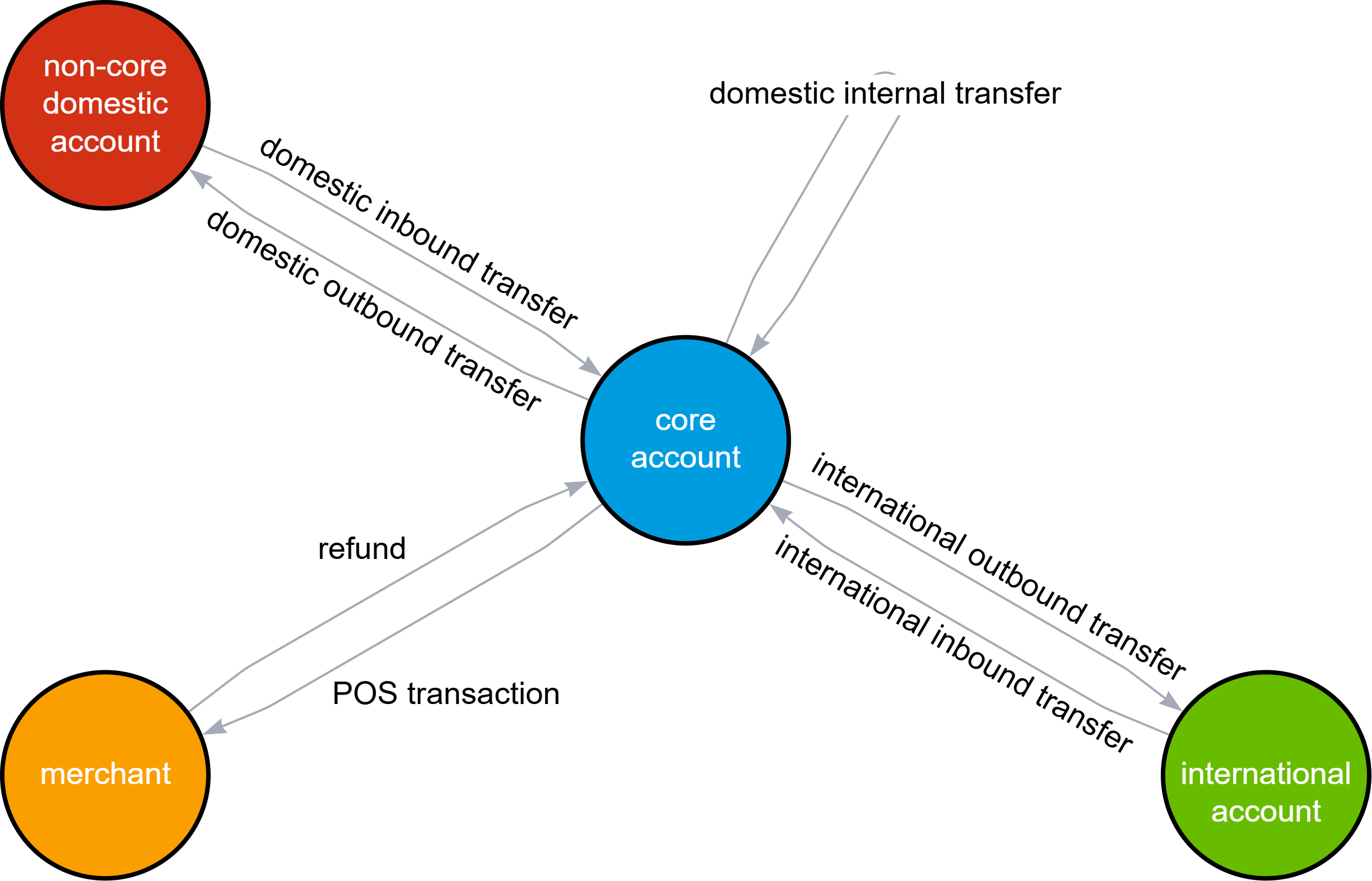}}
\caption{Schema of the graph showing the node and edge types.}
\label{graph_schema}
\end{figure}

To construct a graphical representation that encapsulates all the interactions between the accounts in the transaction data, we define a graph schema that incorporates the maximum available information that can be represented through a transaction network of accounts. Fig.~\ref{graph_schema} outlines the design of the population that is included on this graph.

The graph includes four types of nodes:

\begin{enumerate}
\item Core account – current accounts with core brands in NatWest retail banking that are based in the UK.
\item Non-core account – a UK-based account that is not a core account and has sent or received a transaction from a core domestic account. This could be any external UK account or non-core internal account in NatWest.
\item Foreign account – a non-UK based account (international accounts).
\item Merchant – any merchant that has received a point of sales (POS) payment or issued a refund to a core account.  
\end{enumerate}

These 4 node types induce 7 types of edges on the graph which can be seen in Fig.~\ref{graph_schema}. While the edge types are described individually, they all indicate the same underlying interaction on the network – a flow of money between two accounts – and are uniquely identifiable using the pair of sender-receiver account types. Following this schema, we include transactions over a week to create the graph for training and inference, containing over 100 million edges and over 10 million nodes. In our experiments, we discovered that a week of transactions reduces the variation in transaction habits seen over shorter periods while preventing too much noise, as observed over longer periods.

\subsection{Embedding Algorithm}\label{GEWG}
To generate embeddings for nodes seen on the transaction graph, we use the GraphSAGE algorithm which leverages local neighbourhood information to learn low-dimensional vector representations of nodes. Our embedding model framework can be broken down into three main components:

\subsubsection{Feature Aggregation}\label{FA}
As the key to GraphSAGE being an inductive algorithm, the first stage computes a weighted aggregate of features from a node's neighbours to generate an embedding for the central node. Following the aggregator functions described by Hamilton et al. \cite{b10} (using mean, max-pooling, and LSTM), we used the mean aggregator in our implementation, which balances computational efficiency and representational power for our transaction graph. All the node embeddings on the graph are randomly initialised before training.

\subsubsection{Neighbourhood Sampling}\label{NS}
To tackle the challenge of achieving a reasonable compute time, GraphSAGE utilises a sampling strategy when choosing the neighbourhood to apply the aggregator function to. While transaction networks are typically sparse, i.e., each node only connects to few other nodes, they often contain some ‘super-connected’ nodes which are connected to hundreds or even thousands of other nodes (e.g., a merchant with many active customers, like a supermarket). GraphSAGE effectively combats the runtime for aggregating these cases by sampling their neighbourhood. Our strategy for neighbourhood sampling was to do this using an, optionally weighted, uniform random node sampler with a customisable hyperparameter, fanout, to choose the number of neighbours sampled per node.

\subsubsection{Loss}\label{LOSS}
Once we have the embedding for a node, we must use a loss function to assess the quality of that embedding. We do this using the unsupervised loss described by Hamilton et al. \cite{b10} which trains the embedding model by maximising similarity in the embeddings of neighbouring nodes and minimising the similarity in the embeddings of non-neighbouring nodes.

This three-stage algorithm enables the model parameters to be trained in mini-batches by using stochastic gradient descent and backpropagation. The output embeddings from our setup are 32-dimensional vectors and the hyperparameters are optimised based on empirical experiments to balance model complexity and performance.

\subsection{Hyperparameters}
Our implementation of GraphSAGE required us to tune the following hyperparameters during training:

\begin{itemize}
\item Embedding dimension – size of the output embedding.
\item Learning rate – for gradient descent.
\item Number of negative samples (for every positive sample) – to populate the negative samples in the unsupervised loss function. Negative samples are random samples that don't share an edge with the center node.
\item Hidden layer size – number of neurons for the hidden layer in the graph neural network.
\item Fanout – the number of positive samples used per node in the dataloader during training.
\item Epochs – number of iterations for the training.
\item Edge sampler probabilities – the use of edge weights for obtaining a weighted distribution to sample the positive samples.
\item Batch size – the size of every mini-batch used in training.
\end{itemize}

Optimising the hyperparameters used in our framework was done in two phases. The original GraphSAGE paper, while proposed as an unsupervised representation learning algorithm, primarily used its performance on supervised learning tasks for evaluation. However, it also evaluated performance based on the value of the loss criteria, which did not require labelled data. This intuitive evaluation of the model performance was suitable for us as training our model was a completely unsupervised task and we did not want to bias our training weights towards any specific prediction task. Initially, we began experimentation by investigating the loss criteria across various model training iterations with a range of hyperparameters.

For each hyperparameter, Table~\ref{HTO} summarises our learnings from the first phase of hyperparameter tuning.

\begin{table}[ht]
\centering
\renewcommand{\arraystretch}{1.2} 

\begin{tabularx}{\linewidth}{|p{2cm}|X|}
\hline
Hyperparameter      & Observations     \\ \hline
Embedding dimension      & We initialised with 16 and found that was insufficient to summarise the amount of information on the graph based on plots generated in the next section. Hence, we decided to go with 32. \\ \hline
Learning rate      & A larger learning rate was causing the gradient descent to overshoot – possibly due to the embedded representations being between 0 and 1.       \\ \hline
Number of negative samples   & We observed a significant improvement to the loss values with higher negative samples.    \\ \hline
Size of the hidden layer & A larger hidden layer size was causing memory issues, and hence we tried to maximise this value.       \\ \hline
Fanout              & Since the negative samples are per positive sample, increasing this was also causing memory issues, so we tried to maximise the number of positive samples as much as possible.          \\ \hline
Epochs              & Training loss was plateauing after 10 epochs in every experiment.       \\ \hline
Edge sampler & We attempted to use transaction amounts and other statistics as edge weights, but the runtime was severely impacted (more than doubled) and hence decided not to use them.   \\ \hline
Batch size         & We maximised this as much as possible to maximise our GPU memory utilisation.   \\ \hline
\end{tabularx}
\caption{Outcomes of hyperparameter tuning from the first phase.}
\label{HTO}
\end{table}

Upon completion of the first phase of hyperparameter tuning, a key observation was the strong impact of changing the number of negative samples on the loss value. It was notable that the loss value almost always decreased when the negative samples were raised. Upon diving deeper, we discovered that since the number of negative samples was used as a multiplier on the loss function when minimising similarity to those samples, raising the number of negative samples would always result in the loss value being lower as long as other hyperparameters remained the same. This was a limitation on improving the performance of our unsupervised GraphSAGE implementation.

To overcome this, we defined a second criterion to evaluate the performance of the model. Our approach was inspired by the loss function but does not use any of the hyperparameters in its calculation. The loss function, on a fundamental level, is a binary cross entropy loss function that treats the dot product similarity between the neighbouring nodes as the positive examples and the dot product similarity between non-neighbouring nodes as the negative examples. To judge whether the model was doing this effectively, we defined a metric to compare the cosine similarity of the final embeddings of nodes that were neighbours against that of nodes that were not neighbours with the expectation of the former being noticeably larger than the latter. The positive examples are obtained by considering all edges on the graph and the negative examples are generated by sampling non-existent edges uniformly at random without repetition. The cosine similarity between the embeddings of two nodes, $u$ and $v$, was calculated using \eqref{cs}.

\begin{equation}
cos(u,v)=\frac{\textbf{z}_u{\cdot}\textbf{z}_v}{\|\textbf{z}_u\|\|\textbf{z}_v\|+\varepsilon}\label{cs}
\end{equation}

where $\textbf{z}_i$ is the embedding for node $i$, and $\varepsilon$ is used to prevent 0 division. Table~\ref{NOP} shows the cosine similarity for the positive and negative examples for the hyperparameters selected over the first round of tuning. Despite achieving the lowest loss value, this configuration of the model was not effectively distinguishing between neighbouring and non-neighbouring nodes. Hence, we continued tuning the hyperparameters based on the cosine similarity metric and obtained the final configuration as shown in Table~\ref{OptHP}. While other hyperparameters showed minor differences, the number of negative samples used had the most significant impact on performance. 

\begin{table}[ht]
\centering
\renewcommand{\arraystretch}{1.2} 
\begin{tabularx}{\columnwidth}{|X|X|X|X|X|}
\hline
Negative samples & Loss Value & Average cosine similarity between neighbouring nodes & Average cosine similarity between non-neighbouring nodes & Difference in average cosine similarity \\ \hline
13 & 0.2008 & 0.0902 & 0.0758 & 0.0204            \\ \hline
\end{tabularx}
\caption{Average cosine similarities for sub-optimal model chosen based on loss value.}
\label{NOP}
\end{table}

\begin{table}[ht]
\centering
\renewcommand{\arraystretch}{1.2} 
\begin{tabularx}{\columnwidth}{|X|X|X|X|X|}
\hline
Negative samples & Loss Value & Average cosine similarity between neighbouring nodes & Average cosine similarity between non-neighbouring nodes & Difference in average cosine similarity \\ \hline
2 & 0.3302 & 0.2416 & 0.0475 & 0.1914            \\ \hline
\end{tabularx}
\caption{New optimal hyperparameters were chosen based on average cosine similarity}
\label{OptHP}
\end{table}
  
\section{Validation and Exploration of Embeddings}
This section expands upon how we validated the quality of the embeddings generated by our framework and attempt to uncover any deeper topological information regarding the data that they may have captured. For this section, we assume that analysis has been conducted on core account nodes only (unless mentioned otherwise) as these are the node type of concern to us. First, we will focus on establishing that the training framework is successfully achieving its goal – to embed the similarity between neighbouring nodes and dissimilarity between non-neighbouring nodes. As defined earlier, we will use the cosine similarity metric to showcase this. In the previous section, we applied this metric to the embeddings generated from the same week of transactions that was used to train the model, whereas here we will apply it to the inferred embeddings of transaction graphs from subsequent weeks. Fig.~\ref{cos_sim} illustrates the cosine similarity for the inferred embeddings of neighbouring and non-neighbouring nodes on graphs over the course of a 10-week period. Each week, a new graph is generated based on the transactions from that week and the trained model is used to perform inference on this graph to obtain node embeddings. The plot demonstrates a clear distinction between the two sets as cosine similarity for neighbouring nodes sits comfortably higher than that for non-neighbouring nodes. It also shows a simple regression line for both sets along with their 95\% confidence interval, which remains distinct across the testing period.

\begin{nolinenumbers}
\begin{figure}[htbp]
\centerline{\includegraphics[scale=0.5]{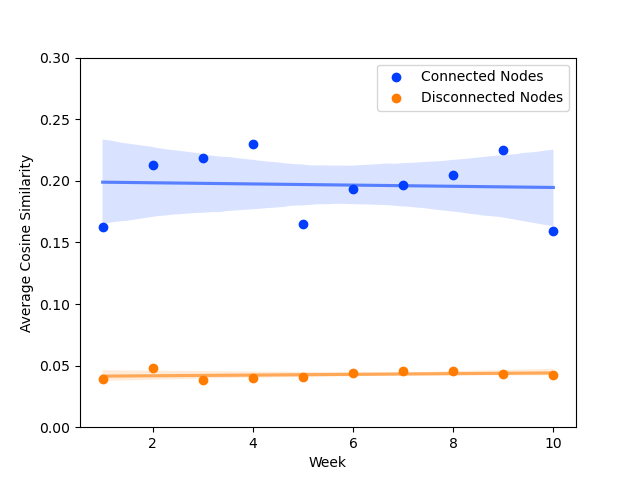}}
\caption{Average cosine similarity between the embeddings of neighbouring and non-neighbouring nodes.}
\label{cos_sim}
\end{figure}
\end{nolinenumbers}

We propose an empirical approach to understand whether the embeddings are able to capture any deeper topological information from the graph. An intuitive way of understanding the embeddings and their distribution is to project them to a lower dimensional space and observe their trends. To do this, we experimented with using t-SNE and UMAP (Uniform Manifold Approximation and Projection) to generate 2-dimensional representations of our embeddings. Unfortunately, t-SNE required significantly larger compute and it was unsustainable to use it to generate the representations for all our experiments, so we decided to focus on using UMAP. We used a standard default configuration UMAP model to map our 32-dimensional core node embeddings to 2 dimensions using which, we hypothesised a few attributes that may be captured by the embeddings and attempted to visualise them on plots. 

\subsection{Geographical Locations}
\begin{figure*}[htbp]
\centerline{\includegraphics[scale=0.5]{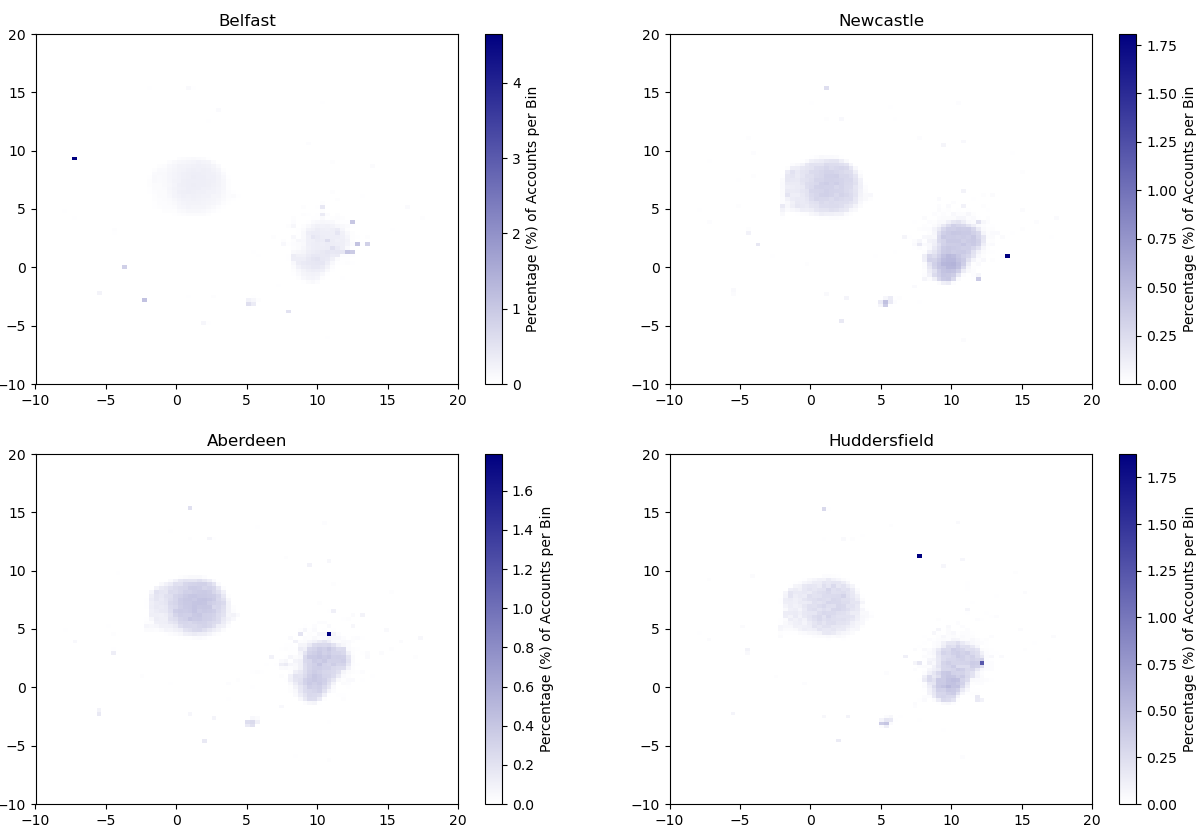}}
\caption{Density plot of UMAP embeddings for different geographical locations.}
\label{geo_plot}
\end{figure*}
Due to the inclusion of merchant nodes on the graph, people in similar geographical vicinities will be represented by the embeddings as they will share some retailers for their day-to-day purchases. Further, this effect will also be amplified as people tend to transfer money to people in the same geographical vicinity as them. Fig.~\ref{geo_plot} shows 4 density plots where the opacity of each pixel is determined by the number of accounts with their UMAP embedding in that region. Specifically, a darker pixel indicates more accounts with an UMAP embedding belonging to that region. The plots have been created by filtering the accounts based on the postcode area of the branch that they are registered with.

In Fig.~\ref{geo_plot}, we showcase the geographical clustering that is induced in the embeddings through our model. Most noticeably, Belfast can be seen clustered densely in a small cluster in the top left corner, containing over 4\% of the total points. This likely results due to a majority of customers banking with Ulster Bank in Northern Ireland. Similar clusters can be observed on a smaller scale for Newcastle, Aberdeen, and Huddersfield, demonstrating the model’s ability to induce geographical properties in the embeddings purely based on an account's transaction activity. 

\subsection{Age Groups}
\begin{figure*}[htbp]
\centerline{\includegraphics[scale=0.5]{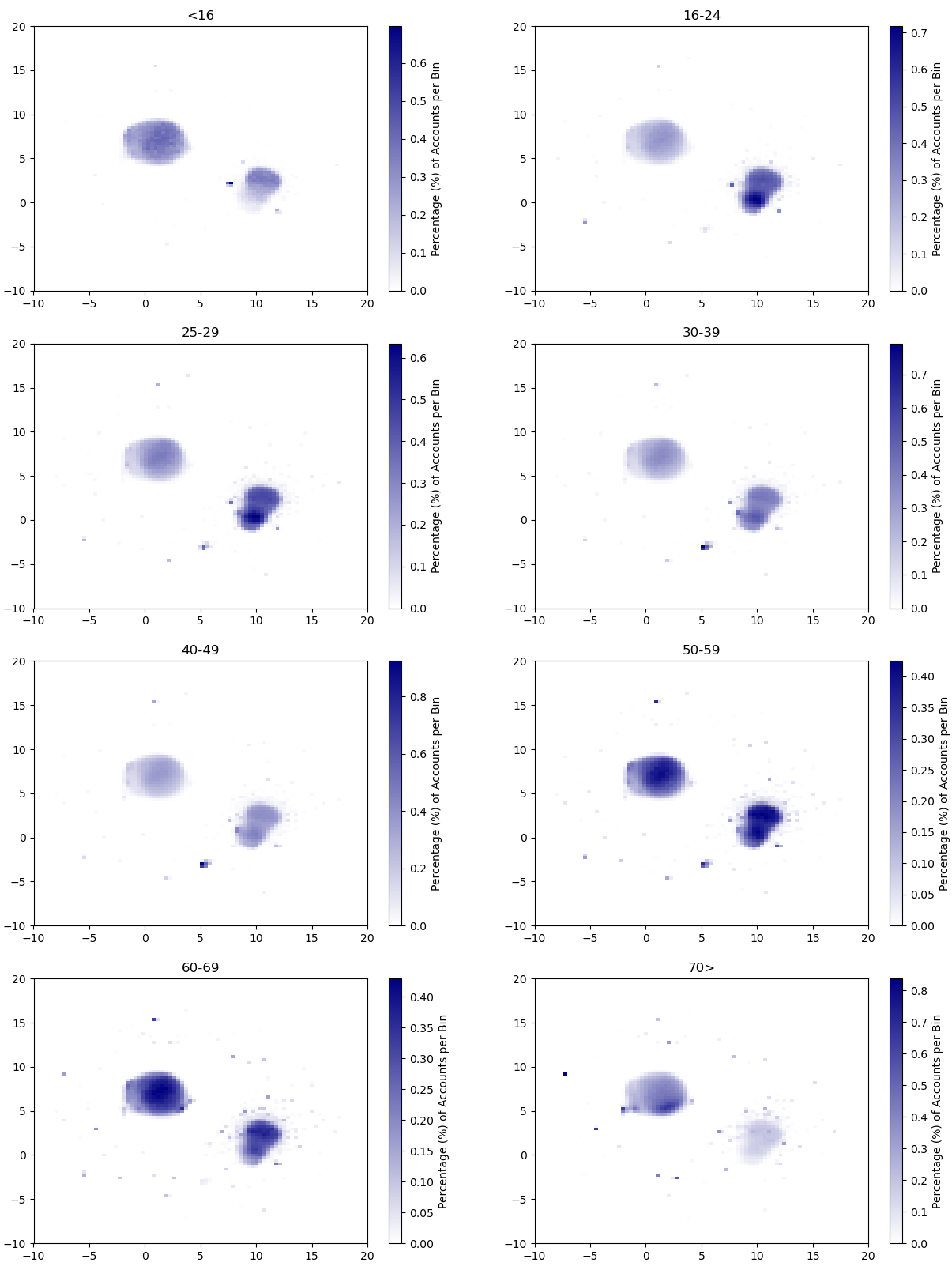}}
\caption{Density plot of UMAP embeddings for different age groups.}
\label{age_plot}
\end{figure*}
Given the holistic view of transaction data contained within the graph, it is possible that the embeddings can capture underlying patterns within the account holders based on their age. If we think of groups of neighbouring nodes as transaction neighbourhoods, it would be fair to say that the transaction neighbourhoods of child, young adult, middle aged and old account holders are likely to be different. Fig.~\ref{age_plot} attempts to discover whether such patterns exist within the UMAP embeddings for different age groups. As described in the previous subsection, this is also a density plot.

Fig.~\ref{age_plot} showcases a couple of different patterns within the UMAP embeddings – firstly, the densest cluster for certain age groups seem to be concentrated in small clusters (as seen in groups with ages $<$16, 30-39 and 40-49) and secondly, generally the densest clusters for different age groups move from bottom right to top left as age increases. This indicates that there are certain differences between these accounts that the embeddings can successfully capture using their transaction activities.

\subsection{Account Types}

\begin{figure}[htbp]
\centerline{\includegraphics[scale=0.3]{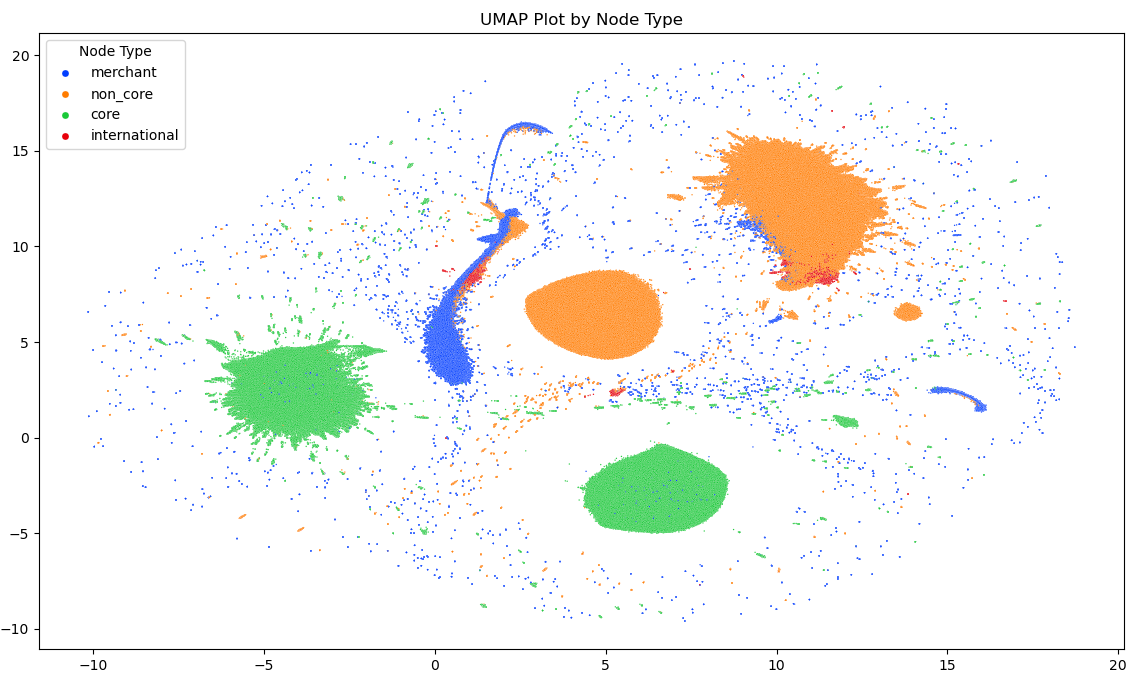}}
\caption{Scatter plot of UMAP embeddings colour-coded by node types.}
\label{nt_plot}
\end{figure}

\begin{figure}[htbp]
\centerline{\includegraphics[scale=0.3]{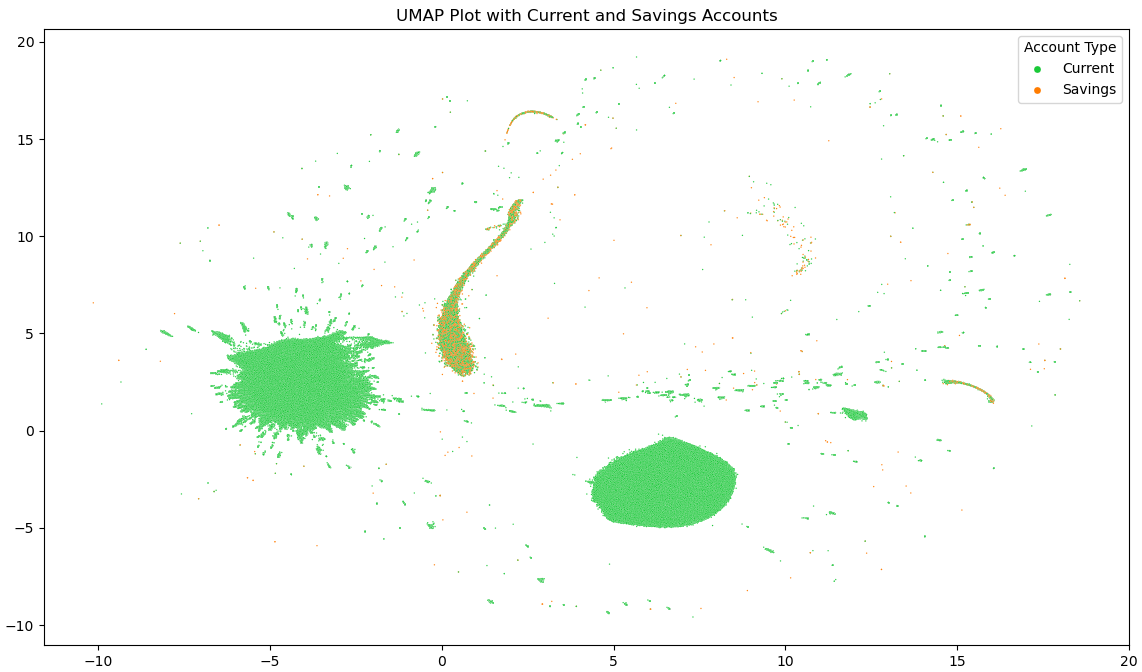}}
\caption{Scatter plot of UMAP embeddings colour-coded by account type.}
\label{cs_plot}
\end{figure}

Since the embeddings for all node types are computed using the same sampling and aggregation strategy, their resulting embeddings belong to the same latent space. To leverage this property, we used a UMAP model to map the embeddings from all node types to 2 dimensions. Upon looking at the scatter plot for their UMAP embeddings, our first observation was that the points were, naturally, clustered based on their node type. Fig.~\ref{nt_plot} shows this phenomenon. To dive further into this, we recalled our population design where NatWest savings accounts were included under the non-core nodes. As savings accounts cannot be used to make merchant transactions and can only transact with their parent accounts, we expected them to have a distinct transaction neighbourhood alongside their respective NatWest current accounts. Upon filtering the UMAP embeddings scatterplot to include only NatWest accounts, we were able to demonstrate this in Fig.~\ref{cs_plot} by shading the points based on their account type.

The previous subsections demonstrate the ability of the model to capture topological information beyond just the connectedness of the graph. This information is induced through the transactional behaviours of each account and can be a highly valuable proxy for account information to any downstream applications. It is also to be noted that UMAP is a dimensionality reduction algorithm and will inevitably not be able to represent all the information contained within the original 32-dimensional embeddings generated by GraphSAGE – i.e., they only provide an estimate of all the information contained in the original embeddings. With that said, they still provide us with a very tangible validation of the successful training of the GraphSAGE model.

\section{Application in Money Mule Detection}
The embeddings generated by a GraphSAGE model can be applied to various scenarios, with one of the most suitable use cases in financial services being money mule detection. Money mules act as intermediaries between suspicious and legitimate accounts and exhibit unique transactional behaviours within the network. Our embeddings capture both local topological patterns and higher-order connectivity in the transaction network, making them effective in representing these behaviours in downstream tasks. 

\subsection{Experimental Setup}
Our graph construction follows the methodology outlined in Section II, modelling bank transfers between current accounts and POS transactions between current accounts and merchants. Transaction data is aggregated weekly to generate dynamic node representations using our framework. To train the downstream fraud detection model, we combine traditional tabular account-level features with fraud labels derived from the bank’s fraud management system, which categorises accounts based on confirmed fraud events. To evaluate the usefulness of graph embeddings, we compare two models: a baseline model trained using only the tabular account-level features and another one trained with the embeddings as additional features. Performance is measured using PR-AUC (area under the precision-recall curve) and precision@k (the precision for the top-k positive predictions), reflecting the model’s ability to prioritise high-risk accounts.

\subsection{Result Discussion}
Table~\ref{CR} summarises the performance gains achieved by the model using the embeddings in comparison to the baseline model:

\begin{table}[ht]
\centering
\renewcommand{\arraystretch}{1.2} 
\begin{tabularx}{\columnwidth}{|X|X|}
\hline
Metrics      & Relative improvement \\ \hline
PR-AUC      & 4.3\%            \\ \hline
Precision@20      & 57.1\%      \\ \hline
Precision@50   & 22.2\%            \\ \hline
Precision@200   & 11.1\%            \\ \hline
\end{tabularx}
\caption{Comparison metrics from downstream classification model.}
\label{CR}
\end{table}

The results demonstrate that graph embeddings significantly enhance the model’s ability to prioritise high-risk accounts in real-world fraud detection workflows. Most notably, precision@20 improves by 57.1\%, indicating that embeddings enable the model to surface structurally suspicious accounts—those embedded in suspicious transaction clusters or ‘hub-and-spoke’ networks— earlier in the ranked predictions. Such accounts often exhibit latent relational patterns (e.g., rapid fund dispersal across previously unconnected entities) that tabular features fail to capture. These improvements align with the operational realities of fraud or financial crime use cases: analysts prioritise investigating top-ranked alerts due to limited bandwidth, and even marginal improvements in early precision massively reduces investigative overhead.

The improvement is less significant for higher values of k (e.g., +11.1\% for precision@200). However, the consistent improvements across all precision@k metrics validate the broader utility of our embeddings. Meanwhile, the modest 4.3\% PR-AUC gain reflects the metric’s sensitivity to class imbalance and its focus on global ranking quality rather than operational thresholds. While PR-AUC stability confirms that embeddings do not degrade overall performance, their true value lies in sharpening precision at the most critical investigative junctures—a distinction that underscores the limitations of aggregate metrics (e.g., PR-AUC) in fraud detection evaluation.

For our money mule detection use case, the graph embeddings provide a mechanistic advantage by embedding transactional relationships and community structures inherent to fraudulent networks. The precision@20 improvement, for instance, suggests that embeddings help identify accounts exhibiting sudden deviations in interaction patterns (e.g., bursts of cross-border transfers) or indirect ties to known mules—signals often obscured in tabular feature spaces. Such capabilities are critical in financial crime, where malicious actors deliberately obfuscate activities across interconnected accounts. In general, the embedding exhibit an ability to consistently improve the precision@k — a metric most aligned with the investigation efficiency for a fraud or financial crime team demonstrates its value as a highly effective feature for prioritising high-risk accounts in a mule detection problem. 

\section{Conclusion}
This paper establishes GraphSAGE as a scalable and adaptable framework for analysing complex transactional networks in retail banking, addressing critical limitations of traditional graph embedding methods. By leveraging GraphSAGE’s inductive capability, we generate embeddings that generalise to unseen nodes, allowing us to infer embeddings over time in the same latent space - a fundamental requirement for financial institutions managing dynamic, real-world transactional data. Our methodology, which constructs a transaction network using anonymised customer transactions and trains a GraphSAGE model to encode relational patterns, reveals interpretable clusters aligned with geographic regions and demographic segments. These clusters validate the embeddings’ ability to capture structural and contextual insights.

The integration of the embeddings into a money mule detection model demonstrates their practical utility, significantly improving the prioritisation of high-risk accounts and showcasing their value in downstream tasks. However, the broader contribution of this work lies in its blueprint for financial institutions to harness graph machine learning to improve their understanding and utilisation of in-house transactional data. GraphSAGE’s scalability ensures compatibility with banking-scale data, while our contributions to the interpretability of the resulting embeddings bridges the gap between technical outputs and actionable business insights.

Our framework enables the inference of embeddings belonging to the same latent space continuously over a long period of time. We demonstrate inferencing embeddings for accounts every week, which gives us the information about ‘where’ the transaction neighbourhood for the account is in that week. As a next step, this work can be progressed further by exploring ways to use the temporal information contained within the evolution of the embeddings, i.e., by addressing how the transaction neighbourhood for the account has changed over its history. Transactional behaviours over time are a notable asset in financial services, e.g., as a contributing factor for detecting fraud, and a temporal study of embeddings like these would be a valuable contribution to the field.



\begin{thebibliography}{14}
\bibitem{b1} Tang, J., Qu, M., Wang, M., Zhang, M., Yan, J., Mei, Q.(2015). ``LINE: Large-scale Information Network Embedding.'' Proceedings of the 24th International Conference on World Wide Web, pp. 1067--1077.
\bibitem{b2} M. Ou, P. Cui, J. Pei, Z. Zhang, and W. Zhu, ``Asymmetric transitivity preserving graph embedding.'' in Proceedings of the 22nd ACM SIGKDD international conference on Knowledge discovery and data mining, 2016, pp. 1105–1114.
\bibitem{b3} U. Goswami, J. Rani, H. Kodamana, P. K. Tamboli, and P. D. Vaswani, ``A graph embedding based fault detection framework for process systems with multi-variate time-series datasets.``, Digital Chemical Engineering, vol. 10, p. 100135, Mar. 2024
\bibitem{b4} Perozzi, B., Al-Rfou, R., Skiena, S. (2014). ``DeepWalk: online learning of social representations.`` Proceedings of the 20th ACM SIGKDD International Conference on Knowledge Discovery and Data Mining, 701–710.
\bibitem{b5} J. V. V. Sriram Sasank, G. R. Sahith, K. Abhinav, and M. Belwal, ``Credit Card Fraud Detection Using Various Classification and Sampling Techniques: A Comparative Study``, in 2019 International Conference on Communication and Electronics Systems (ICCES), Jul. 2019, pp. 1713–1718
\bibitem{b6} R. Van Belle, B. Baesens, and J. De Weerdt, ``CATCHM: A novel network-based credit card fraud detection method using node representation learning``, Decision Support Systems, vol. 164, p. 113866, Jan. 2023.
\bibitem{b7} Grover, A., Leskovec, J. (2016). ``node2vec: Scalable feature learning for networks,'' Proceedings of the 22nd ACM SIGKDD International Conference on Knowledge Discovery and Data Mining, 855–864.
\bibitem{b8} Kipf, T. N., Welling, M. (2017). ``Semi-Supervised Classification with Graph Convolutional Networks,'' Proceedings of the 5th International Conference on Learning Representations.
\bibitem{b9} Veličković, P., Cucurull, G., Casanova, A., Romero, A., Liò, P., Bengio, Y. (2017). ``Graph Attention Network,'' ICLR 2018.
\bibitem{b10} Hamilton, W., Ying, Z., Leskovec, J. (2017). ``Inductive Representation Learning on Large Graphs,'' Advances in Neural Information Processing Systems, 11.
\bibitem{b11} C. B. Bruss, A. Khazane, J. Rider, R. Serpe, A. Gogoglou, and K. E. Hines, ``DeepTrax: Embedding Graphs of Financial Transactions,'' 2019. [Online]. Available: https://arxiv.org/abs/1907.07225
\bibitem{b12} Y. Dong, N. V. Chawla, and A. Swami, ``metapath2vec: Scalable Representation Learning for Heterogeneous Networks``, in Proceedings of the 23rd ACM SIGKDD International Conference on Knowledge Discovery and Data Mining, in KDD ’17. New York, NY, USA: Association for Computing Machinery, Aug. 2017, pp. 135–144.
\bibitem{b13} Van Belle, R., Van Damme, C., Tytgat, H. and De Weerdt, J., 2022. ``Inductive graph representation learning for fraud detection''. Expert Systems with Applications, 193, p.116463.
\end{thebibliography}
\end{document}